\documentclass{article}

\usepackage{cite} 
\usepackage[english]{babel}
\usepackage{titlesec}
\usepackage{xcolor}%
\usepackage{textcomp}%
\usepackage{manyfoot}%
\usepackage{booktabs}%
\usepackage{algorithm}%
\usepackage{algorithmicx}%
\usepackage{algpseudocode}%
\usepackage{listings}%
\usepackage{amsmath,amssymb,amsfonts}
\usepackage{multirow}
\usepackage[letterpaper,top=2cm,bottom=2cm,left=3cm,right=3cm,marginparwidth=1.75cm]{geometry}
\usepackage{graphicx}
\usepackage[colorlinks=true, allcolors=blue]{hyperref}
\usepackage{tikz}
\usepackage{subcaption}
\usepackage{authblk}
\usepackage[title]{appendix}
\usepackage{tabularx} 

\title{Adaptive Locally Linear Embedding}
\author[1]{Ali Goli}
\author[1]{Mahdieh Alizadeh}
\author[1,2]{{\normalsize Hadi Sadoghi Yazdi}\thanks{Corresponding author: h-sadoghi@um.ac.ir}}

\affil[1]{\normalsize Department of Computer Engineering, Ferdowsi University of Mashhad, Mashhad, Iran}
\affil[2]{\normalsize Center of Excellence in Soft Computing and Intelligent Information Processing, Ferdowsi University of Mashhad, Mashhad, Iran}

\begin{document}
\maketitle

\begin{abstract}
Manifold learning techniques, such as Locally linear embedding (LLE), are designed to preserve the local neighborhood structures of high-dimensional data during dimensionality reduction. Traditional LLE employs Euclidean distance to define neighborhoods, which can struggle to capture the intrinsic geometric relationships within complex data. A novel approach, Adaptive locally linear embedding(ALLE), is introduced to address this limitation by incorporating a dynamic, data-driven metric that enhances topological preservation. This method redefines the concept of proximity by focusing on topological neighborhood inclusion rather than fixed distances. By adapting the metric based on the local structure of the data, it achieves superior neighborhood preservation, particularly for datasets with complex geometries and high-dimensional structures. Experimental results demonstrate that ALLE significantly improves the alignment between neighborhoods in the input and feature spaces, resulting in more accurate and topologically faithful embeddings. This approach advances manifold learning by tailoring distance metrics to the underlying data, providing a robust solution for capturing intricate relationships in high-dimensional datasets.
\end{abstract}

\providecommand{\keywords}[1]
{
  \small	
  \textbf{\textit{Keywords---}} #1
}

\section{Introduction}

Locally linear embedding(LLE) is a prominent manifold learning technique designed to reduce the dimensionality of high-dimensional datasets while preserving their intrinsic geometric structure. Proposed by Roweis and Saul, LLE operates through a systematic process that includes identifying the K-nearest neighbors for each data point, calculating reconstruction weights to express each point as a linear combination of its neighbors, and ultimately generating a low-dimensional representation that retains local relationships \cite{roweis2000lle}. However, LLE traditionally relies on fixed distance metrics, such as Euclidean distance, which may inadequately represent complex data distributions and fail to capture nuanced topological relationships.

In response to these limitations, we introduce a novel approach termed Adaptive LLE(ALLE), which integrates a flexible, data-driven metric into the LLE framework. This method not only learns the metric concurrently with the embedding process but also adapts to the local structure of the data. By employing an adaptive metric, our approach aims to enhance the preservation of neighborhood structures, leading to embeddings that more accurately reflect the intricate relationships present in high-dimensional datasets.

The benefits of training a metric alongside LLE include improved neighborhood preservation and more robust embeddings, as the learned metric can dynamically adjust to the local geometry of the data. However, a potential drawback is that without adequate training, the metric may not capture the true data structure, leading to suboptimal embeddings.

Previous studies have explored similar concepts, integrating metrics into manifold learning frameworks. For instance, approaches such as RLLE and Kernel LLE have aimed to enhance LLE's robustness and capacity to capture non-linear structures \cite{chang2006robust,zhang2007kernel}. These methodologies typically adhere to specific topological structures that facilitate the preservation of local neighborhoods. However, our proposed method distinguishes itself by emphasizing the adaptability of the metric based on data-driven principles, thereby offering a more generalized framework for neighborhood preservation.

In this paper, we aim to further develop this concept, providing a detailed account of our methodology and its theoretical underpinnings, while also comparing it against existing techniques in terms of topology preservation and neighborhood accuracy. Through this investigation, we seek to contribute to the ongoing discourse on manifold learning and metric learning, highlighting the critical interplay between these domains in advancing dimensionality reduction techniques. The implementation of our proposed methods, along with supplementary materials, is available at \href{https://github.com/goli3148/MetricLLE.git}{GitHub repository}.

\begin{figure}[t]
    \centering
    \begin{subfigure}{0.48\textwidth}
        \centering
        \includegraphics[width=\textwidth]{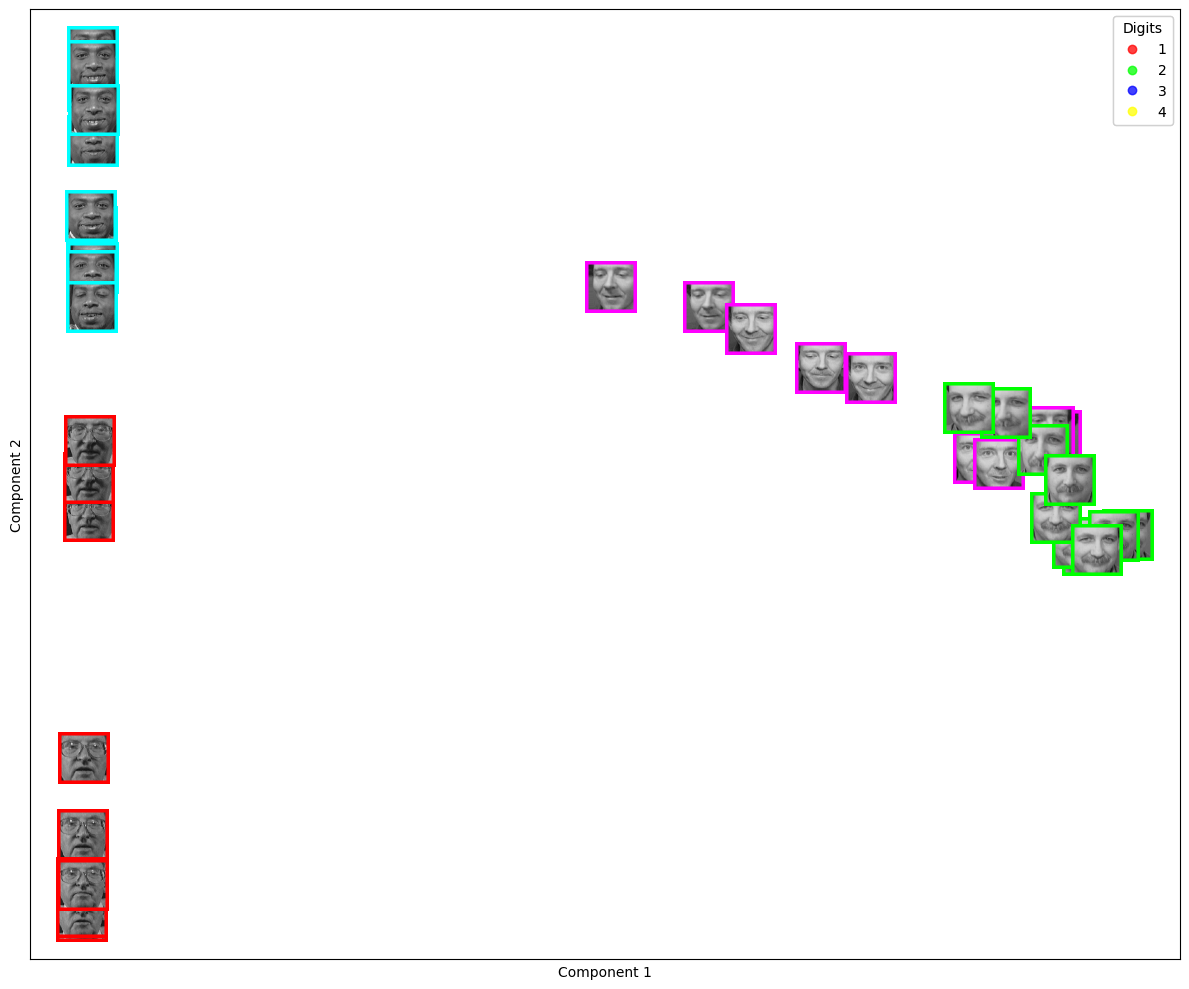}
        \caption{Standard LLE: Entangled embedding due to the inability to simplify complex regions.}
        \label{fig:motivation-standard}
    \end{subfigure}
    \hfill
    \begin{subfigure}{0.48\textwidth}
        \centering
        \includegraphics[width=\textwidth]{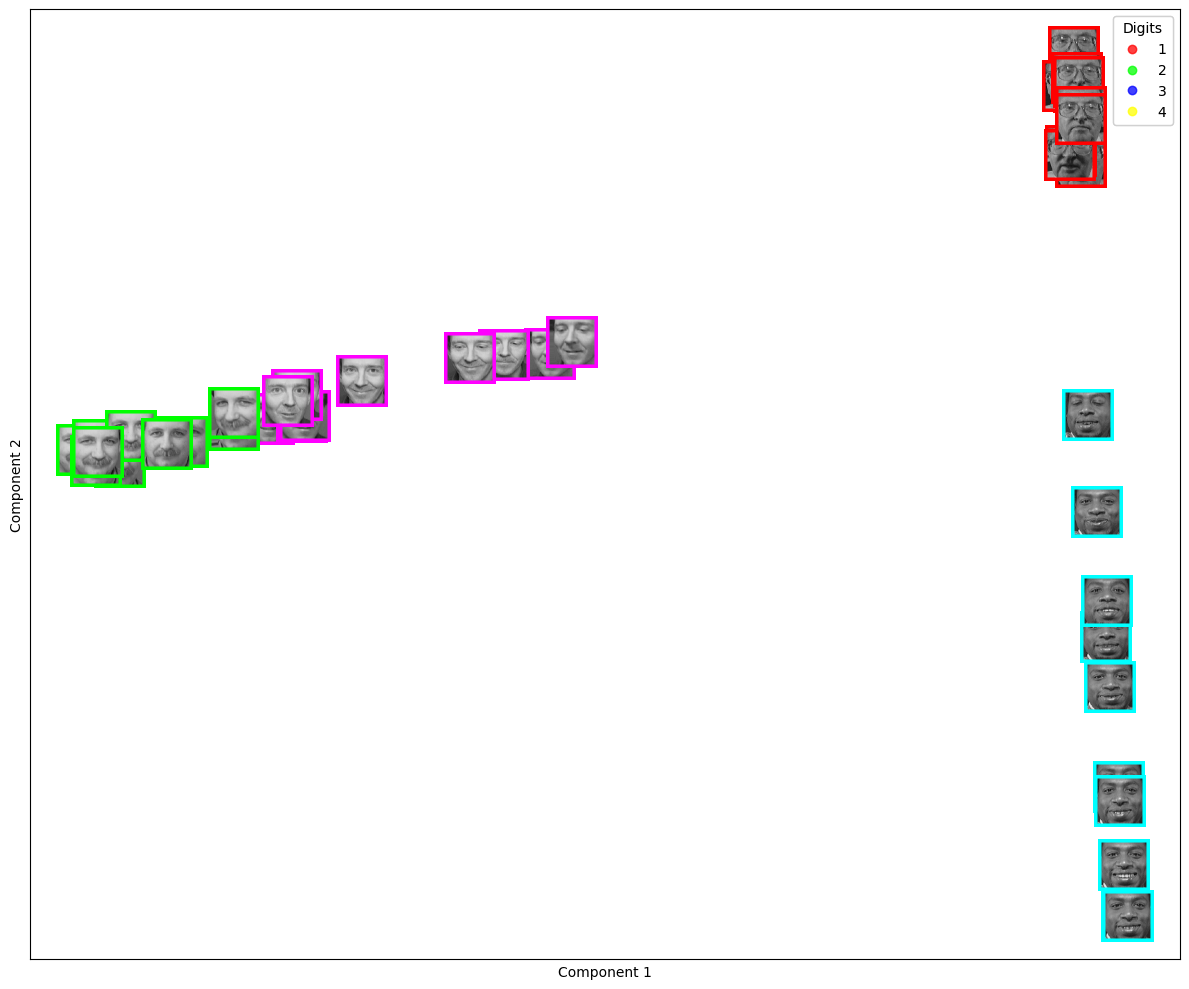}
        \caption{Adaptive LLE: Improved separation of data through metric learning in topology space.}
        \label{fig:motivation-adaptive}
    \end{subfigure}
    \vskip\baselineskip
    \begin{subfigure}{0.48\textwidth}
        \centering
        \includegraphics[width=\textwidth]{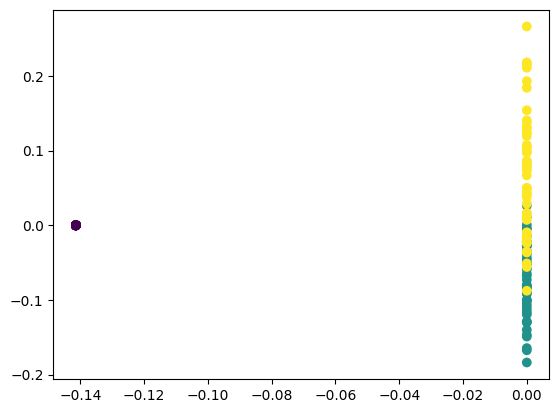}
        \caption{Standard LLE: Entangled embedding due to the inability to simplify complex regions.}
        \label{fig:motivation-standard2}
    \end{subfigure}
    \hfill
    \begin{subfigure}{0.48\textwidth}
        \centering
        \includegraphics[width=\textwidth]{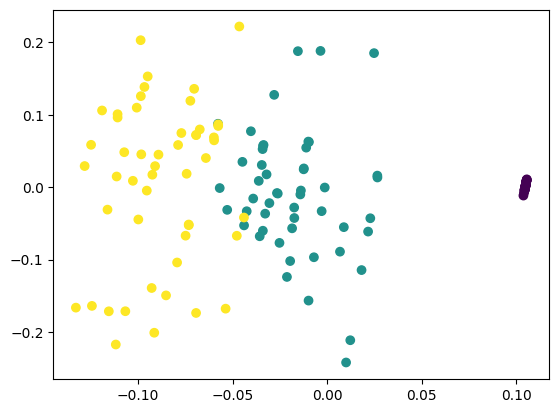}
        \caption{Adaptive LLE: Improved separation of data through metric learning in topology space.}
        \label{fig:motivation-adaptive2}
    \end{subfigure}
    \caption{Comparison of standard LLE and adaptive LLE in handling complex data structures.}
    \label{fig:motivation}
\end{figure}
\subsection{Motivation}
In LLE, the goal is to reduce the dimensionality of data while preserving its local geometric properties. However, the standard approach often struggles with complex data structures, leading to entangled embeddings that fail to separate certain regions effectively, as illustrated in Fig. \ref{fig:motivation-standard}. the proposed method learns a metric in the topology space, which allows for bending, stretching, and compressing the data manifold in ways that better align with the underlying data complexity. 

This adaptive metric enables the deformation of the topology, such that complex regions of the data space are simplified, making it easier to distinguish and separate key features. As shown in Fig. \ref{fig:motivation-adaptive}, proposed approach, produces embeddings with significantly improved separation and alignment compared to the standard method. By reducing the complexity of the data manifold, the proposed method not only disentangles the output but also enhances the accuracy of downstream tasks such as classification.

As depicted in Fig. \ref{fig:motivation-standard2} and Fig. \ref{fig:motivation-adaptive2}, the plots represent a low-dimensional embedding of the Iris dataset \cite{iris_53} obtained using standard LLE and ALLE, respectively. In Fig. \ref{fig:motivation-standard2}, the 2D representation generated by standard LLE demonstrates that the embeddings of two specific classes remain entangled, making it challenging to distinguish between them in the reduced-dimensional space.

In contrast, Fig. \ref{fig:motivation-adaptive2} illustrates the results of applying ALLE, where the topology of the data space is adaptively stretched. This transformation significantly improves the separability of the classes, effectively resolving the entanglement issue observed in the standard LLE embedding. The adaptive approach of ALLE, which leverages metric learning to modify the data's intrinsic geometry, ensures a more discriminative low-dimensional representation while preserving local relationships.

\subsection{Topological Space and Neighborhoods}
In topology, the concept of neighborhoods defines the idea of "closeness" without relying on specific metrics. A neighborhood includes a point and all other points considered "close" by inclusion within an open set, rather than a predefined distance. This abstract notion of proximity offers a powerful flexibility as it emphasizes relationships between points in a more generalized manner, independent of fixed metrics.

A topology on a set \( X \) is a collection \( \mathcal{T} \) of subsets of \( X \), called open sets, that satisfies three conditions: the empty set and the entire set \( X \) are in \( \mathcal{T} \); the union of any collection of sets in \( \mathcal{T} \) is also in \( \mathcal{T} \); and the intersection of any finite collection of sets in \( \mathcal{T} \) is also in \( \mathcal{T} \) \cite{munkres2000topology}.

A neighborhood of a point \( x \) in a topological space \( X \) is a set \( N \) containing \( x \) such that there exists an open set \( O \) where \( x \in O \subseteq N \). In this context, \( N \) contains an open set surrounding \( x \), reflecting a topological sense of closeness \cite{willard2004general}.

\subsection{The Metric Topology}
This section explores one of the fundamental methods of defining a topology on a set, using a metric. A metric introduces the concept of distance between elements of a set, offering a precise way to understand how "close" or "far apart" these elements are. Topologies derived from metrics serve as the cornerstone of modern analysis and provide an intuitive yet rigorous framework for studying continuity and convergence.

Formally, a metric on a set \( X \) is a function \( d: X \times X \to \mathbb{R} \) that assigns a non-negative real number to every pair of points \( x \) and \( y \) in \( X \), representing the distance between them. To qualify as a metric, the function \( d \) must satisfy three conditions. The first is non-negativity, which requires that \( d(x, y) \geq 0 \), with equality only if \( x = y \), ensuring that distinct points always have a positive distance. The second is symmetry, meaning \( d(x, y) = d(y, x) \), reflecting the intuitive idea that the distance between two points is independent of their order. Finally, the triangle inequality states that \( d(x, z) \leq d(x, y) + d(y, z) \), ensuring that the direct distance between two points is never greater than the sum of the distances when traveling through a third point.

Using a metric, the notion of an \(\epsilon\)-ball, denoted \( B_d(x, \epsilon) \), is defined as the set of all points \( y \) in \( X \) such that \( d(x, y) < \epsilon \). This \(\epsilon\)-ball can be thought of as a neighborhood around \( x \), containing all points within a specified radius \( \epsilon \). By considering these neighborhoods for every point in \( X \), a metric topology emerges. This topology organizes the points of \( X \) into a structure that reflects their proximity, as determined by the metric, providing a natural way to study the relationships between points based on ``closeness.''
\cite{munkres2000topology}

\subsection{Innovation: Need for a Flexible Metric}
Traditional metrics, such as Euclidean distance, often fail to adequately represent complex data distributions, particularly in high-dimensional or non-linear datasets. A static metric like Euclidean distance assumes uniform scaling across the data, which leads to inaccuracies in neighborhood preservation. To address this limitation, we propose a data-dependent metric defined as:
\[
d_M(x_i, x_j) = \sqrt{(x_i - x_j)^T M (x_i - x_j)},
\]
where \( M \) is a positive semi-definite matrix learned from the data. This metric is designed to adapt to the local geometry, providing a more flexible and accurate representation of proximity by reflecting both global and local characteristics of the data distribution.
We introduce a novel approach to learning a metric \( d_M(x_i, x_j) \) that adapts to the local geometry of the data, ensuring that neighborhood structures are accurately preserved in the feature space. This metric is learned through an iterative optimization process that minimizes the least-squares error of neighborhood reconstruction, resulting in a more faithful representation of distances in both the input and embedded spaces.
As illustrated in Fig. ~\ref{fig:proposed_method}, our method introduces a data-dependent metric that better preserves neighborhood structures in the input space. By letting the data inform the metric, we achieve embeddings that better reflect the intrinsic relationships within complex, high-dimensional datasets. This adaptive, context-sensitive approach represents a significant advancement in manifold learning, leading to more meaningful embeddings and a more accurate capture of the relationships between data points.

\subsection{Manifold Perspective}
A manifold is a topological space that locally resembles Euclidean space, even though its global structure may be complex. Manifold learning techniques leverage this local Euclidean structure to better understand the behavior of high-dimensional data. The challenge lies in preserving the essential geometric properties of the manifold when embedding it in a lower-dimensional space.
In manifold learning, methods LLE aim to preserve local neighborhoods when mapping high-dimensional data to a lower-dimensional space. However, Euclidean distance, which is typically used to define these neighborhoods, may not faithfully represent the true underlying structure due to its limitations in scaling uniformly across all points. Our learned metric enhances this process by allowing proximity to be defined in a way that is sensitive to local variations in the data, thereby improving the accuracy of neighborhood preservation.

\section{Related Work}

Dimensionality reduction has been a core technique in machine learning and data analysis, with methods such as Principal Component Analysis \cite{hotelling1933analysis} and Multidimensional Scaling \cite{torgerson1952multidimensional} serving as traditional linear embedding approaches. These methods focus on preserving global structures but fail to capture the non-linear relationships inherent in complex data. This limitation has motivated the development of manifold learning techniques.

Tenenbaum et al. \cite{tenenbaum2000global} introduced a global geometric framework for nonlinear dimensionality reduction, capable of preserving the intrinsic geometry of data. Their approach efficiently uncovers nonlinear structures, such as manifolds, using local metric information to infer global relationships.

One of the most prominent methods is LLE, introduced by Roweis and Saul \cite{roweis2000lle}. LLE constructs a low-dimensional representation by preserving local linear relationships, assuming that data lies on or near a non-linear manifold. It maps high-dimensional data to a lower-dimensional space while preserving local geometries. 

Neighborhood selection, which is the only nonlinear step of the LLE, plays a very important role in the performance of the algorithm. If the points are unevenly or sparsely sampled or
contaminated by noise, the estimated reconstruction weights of its neighbors cannot reflect well the local geometry of the manifold in the embedding space, even leading to a large bias
to the embedding result. Most extensions of LLE are concerning the neighborhood selection\cite{Chen2011}. also, LLE uses Euclidean distance to define neighborhoods, which can misrepresent the true geometric relationships in more complex or non-uniformly sampled datasets.

Several enhancements to LLE have been proposed to address its limitations. Pan et al. \cite{PAN2009798} introduced Weighted Locally Linear Embedding (WLLE), which incorporates a novel weighted distance measure to improve neighbor selection and reduce redundancy. WLLE addresses challenges posed by noisy, sparse, or weakly connected data by adapting to the local data distribution. The method demonstrates superior manifold learning performance and robustness to parameter changes compared to standard LLE and NLE. Additionally, WLLE's practical applications, such as in face recognition, highlight its ability to outperform traditional methods like KPCA and kernel direct discriminant analysis in certain scenarios.

\cite{Guihua2008} proposed Kernel Relative Transformation and Relative Transformation to enhance LLE. These transformations aim to improve neighborhood graph construction by reducing the impact of noise, sparsity, and short-circuit edges, ensuring better preservation of the intrinsic structure of data manifolds. 

Chang and Yeung \cite{chang2006robust} introduced RLLE, which incorporates robust statistics to handle outliers effectively, ensuring stable embeddings even in the presence of noisy data. Kernel LLE \cite{zhang2007kernel} extends the original method by introducing non-linear kernel functions, allowing the method to better capture the structure of complex non-linear manifolds. Further advancing LLE's capabilities, \cite{varini2006isolle} proposed Isometric LLE (ISO LLE), which explicitly enforces geodesic distance preservation during embedding. Unlike standard LLE, which focuses on local linear reconstructions, ISO LLE optimizes for global isometry by incorporating pairwise distance constraints, bridging the gap between LLE's local emphasis and Isomap's global geodesic approach. These approaches have improved the applicability of LLE but remain reliant on predefined distance metrics that may not fully reflect the complexity of the data.

Alternative manifold learning methods, such as Isomap \cite{tenenbaum2000global} and Laplacian Eigenmaps \cite{belkin2003laplacian}, aim to address some of these limitations by incorporating geodesic distances and graph-based techniques to preserve both local and global structures. While effective, these methods still face challenges in preserving true neighborhood relationships, particularly in highly complex or sparsely sampled datasets.

To overcome the limitations of fixed distance metrics, metric learning has been introduced as a powerful approach for dynamically adapting the distance function based on the data. Xing et al. \cite{xing2003distance} proposed a Mahalanobis distance-based method that learns a metric to better reflect the geometry of the data, improving clustering performance. Weinberger and Saul's Large Margin Nearest Neighbor algorithm \cite{weinberger2009distance} further demonstrated the effectiveness of adaptive metrics by optimizing neighborhood relationships, achieving significant improvements in classification tasks.

Building on this foundation, recent works have explored the integration of metric learning into manifold learning techniques. Xu and Li \cite{xu2024robust} introduced Robust Locally Nonlinear Embedding, which constructs nonlinear relationships between local neighborhoods using basis functions and constrained least squares. This method aligns neighborhoods with locally nonlinear patches of the manifold, enhancing robustness to noise and outperforming conventional LLE-based methods. Liu et al. \cite{liu2024bearing} proposed the LMR-LLE algorithm, which improves traditional LLE by incorporating multiple neighborhood structures through local mutual representation. This approach fuses diverse local structures into a global reconstruction model, achieving superior performance in tasks such as bearing fault diagnosis.

Recent applications have further demonstrated the versatility of LLE-based methods. Tao et al. \cite{tao2024privacy} proposed a privacy-preserving outsourcing scheme for face recognition, utilizing LLE with sparse orthogonal matrices to reduce computational complexity. While the method enhances efficiency through batch outsourcing and result verification, it faces limitations related to communication latency and scalability. Wang et al. \cite{wang2024spectral} focused on improving graph structure learning with a non-negative orthogonal constraint. Their method enhances clustering accuracy by minimizing intra-class distances and optimizing inter-class topology through category anchors, though challenges related to scalability and initialization sensitivity remain.

The integration of metric learning into manifold learning offers a significant advantage by allowing the distance function to adapt dynamically to the local geometry of the data. This flexibility enables better preservation of neighborhood relationships and improves embedding quality. Our proposed method builds on this foundation by introducing a data-driven metric that adjusts dynamically to the structure of the data, leading to more accurate and robust low-dimensional representations.

\begin{figure}
    \centering
    \includegraphics[width=1\textwidth]{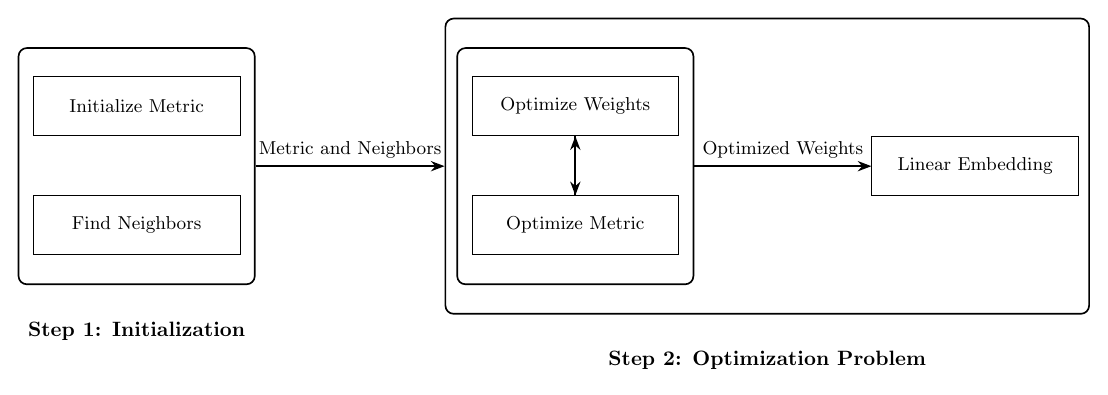}
    \caption{Overview of the proposed method.}
    \label{fig:proposed_method}
\end{figure}
\section{The proposed method}
In the realm of high-dimensional data analysis, traditional metrics like Euclidean distance often fall short, particularly when grappling with complex, non-linear datasets. The inherent assumption of uniform scaling in static metrics can distort neighborhood relationships, leading to misrepresentation in proximity calculations. To overcome these challenges, we introduce a data-dependent metric that adapts to the intricate local geometry of the data. By learning a positive semi-definite matrix \( M \) from the data, our approach not only enhances the fidelity of distance measurements but also preserves essential neighborhood structures, paving the way for more meaningful insights in manifold learning.

As illustrated in Fig.\ref{fig:proposed_method}, the process begins by initializing a metric matrix \( M \) that adapts to the unique structure of the data. This matrix is learned iteratively, starting with an initial guess that is refined through optimization. The method involves finding the \( K \)-nearest neighbors in the high-dimensional input space, guided by the learned metric \( M \), rather than static distances. This step helps to preserve essential topological features by capturing neighborhood relationships that reflect the intrinsic geometry of the data.

After identifying neighbors, we proceed to the linear reconstruction phase, where each data point is expressed as a weighted linear combination of its neighbors. These weights are iteratively optimized to minimize reconstruction error while maintaining the local topology. In parallel, the metric \( M \) is updated based on the reconstruction error, allowing it to better represent the data's manifold structure in subsequent iterations. This alternating optimization loop between linear reconstruction and metric refinement ensures that both the neighborhood preservation and the accuracy of distance measurements are continually improved.

Finally, the optimized metric and reconstruction weights are used to generate the low-dimensional embedding, preserving the manifold's structure as faithfully as possible. This adaptive approach, as detailed in Fig. \ref{fig:proposed_method}, not only enhances the fidelity of distance measurements but also aligns them with the underlying data distribution, enabling more meaningful insights in manifold learning.

\subsection{Notations and definitions}
\label{subsec:notations_definitions}
In this section, we introduce the key notations and definitions that will be used throughout the paper. Let \( \mathcal{X} = \{\mathbf{x}_1, \mathbf{x}_2, \ldots, \mathbf{x}_n\} \subset \mathbb{R}^D \) represent a set of \( n \) data points in a \( D \)-dimensional space. Each data point \( \mathbf{x}_i \in \mathbb{R}^D \) corresponds to a high-dimensional vector. In the context of LLE, let \( K \) denote the number of nearest neighbors considered for each data point. For a given point \( \mathbf{x}_i \), its set of KNN is denoted as \( \mathcal{N}(\mathbf{x}_i) \).

The reconstruction process in LLE assigns weights to neighbors, where \( \mathbf{W} \in \mathbb{R}^{n \times k} \) is the matrix of reconstruction weights. Each entry \( w_{ij} \) represents the weight assigned to the \( j \)-th neighbor of the point \( \mathbf{x}_i \). A positive semi-definite matrix \( \mathbf{M} \) is introduced to represent the learned distance metric. Using \( \mathbf{M} \), the Mahalanobis distance between two points \( \mathbf{x}_i \) and \( \mathbf{x}_j \) is defined as \( d_\mathbf{M}(\mathbf{x}_i, \mathbf{x}_j) = \sqrt{(\mathbf{x}_i - \mathbf{x}_j)^T \mathbf{M} (\mathbf{x}_i - \mathbf{x}_j)} \).

In the dimensionality reduction process, each point \( \mathbf{x}_i \) is mapped to a low-dimensional representation \( \mathbf{y}_i \in \mathbb{R}^d \), where \( d \) is the dimension of the embedding space. The quality of the reconstruction in the high-dimensional space is quantified using the reconstruction error, denoted as \( E(\mathbf{W}) \). This error is given by 
\[
E(\mathbf{W}) = \sum_{i=1}^{n} \left\| \mathbf{x}_i - \sum_{j \in \mathcal{N}(i)} w_{ij} \mathbf{x}_j \right\|_\mathbf{M}^2,
\]
where \( \| \cdot \|_\mathbf{M} \) is the norm induced by the learned metric \( \mathbf{M} \). For each point \( \mathbf{x}_i \), the reconstruction error vector is defined as \( \mathbf{r}_i = \mathbf{x}_i - \sum_{j \in \mathcal{N}(i)} w_{ij} \mathbf{x}_j \), capturing the deviation of \( \mathbf{x}_i \) from its weighted reconstruction based on its neighbors.
These notations and definitions form the basis for the subsequent discussions on the proposed modifications and optimization processes in LLE.

\subsection{Problem Formulation}
\label{subsec:problem_formulation}
In the classical LLE algorithm proposed by Roweis and Saul \cite{roweis2000lle}, the goal is to reduce the dimensionality of a dataset \cite{ghodsi2006dimensionality} while preserving the local neighborhood structure of the data through three key steps. Given a set of points \( \{\mathbf{x}_1, \mathbf{x}_2, \dots, \mathbf{x}_n\} \in \mathbb{R}^D \), the algorithm first identifies the KNN of each data point \cite{bishop2006pattern}. Once the neighbors are determined, LLE computes a set of reconstruction weights that express each point \( x_i \) as a linear combination of its neighbors in the original high-dimensional space. These weights are then utilized to construct a low-dimensional embedding \( \{\mathbf{y_1}, \mathbf{y_2}, \dots, \mathbf{y_N} \} \in \mathbb{R}^d \) (with \( d \ll D \)), ensuring that each point \( y_i \) is similarly represented as a weighted combination of its neighbors, thereby preserving the local structure of the data in the lower-dimensional representation.
while the 
The standard LLE algorithm determines neighboring points using a fixed distance metric, typically the Euclidean distance:
\[
d(\mathbf{x_i}, \mathbf{x_j}) = \|\mathbf{x_i} - \mathbf{x_j}\|_2.
\]
This fixed notion of distance may fail to accurately capture the true local geometry of the data, particularly in cases where the manifold is curved or the data is anisotropically distributed. To address this limitation, we propose a controllable LLE framework in which the distance between points is governed by a learned metric. By utilizing a Mahalanobis distance metric in both the K-nearest neighbors search and the linear reconstruction steps, our approach enhances the algorithm's capacity to adaptively model the local structure of the data. This methodology facilitates a more precise representation of the underlying manifold, as will be discussed further.
\begin{figure}
    \centering
    \includegraphics[width=0.5\textwidth]{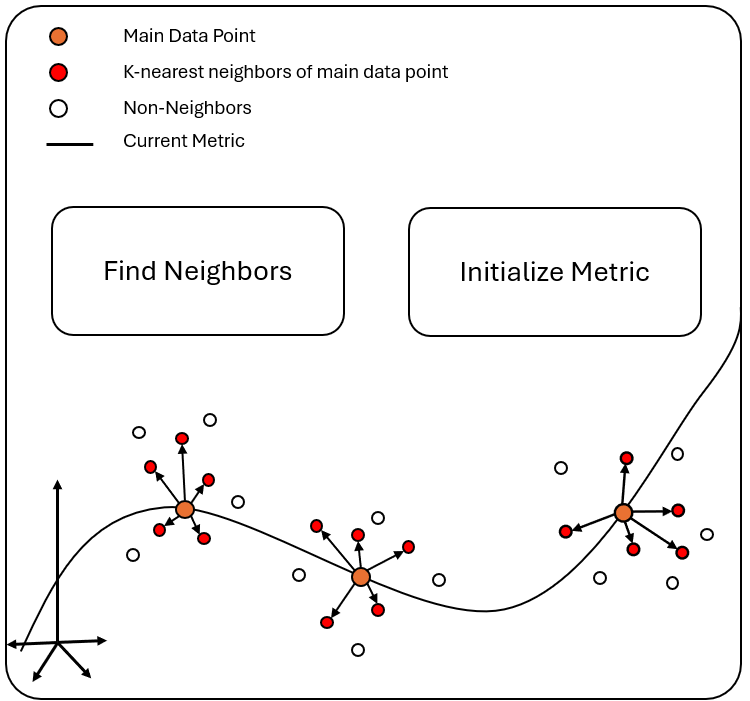}
    \caption{Initialization}
    \label{fig:initialization}
\end{figure}
\subsection{Initialization}
As illustrated int Fig.\ref{fig:initialization}, this phase is a critical preparatory step it involves two sub-steps:
\subsubsection{Find Nearest Neighbors}

In this step, the nearest neighbors of each data point in the high-dimensional space are identified, forming the local neighborhood that is preserved during the dimensionality reduction process. To find the neighbors, a distance metric is employed to measure proximity between points, with the Euclidean distance being the most commonly used, although alternative metrics can be applied depending on the application. For each data point \( x_i \), the distances to all other points are computed, and the \( K \) closest points are selected as its neighbors. These neighborhoods are then stored for use in subsequent steps, such as computing reconstruction weights and embeddings, ensuring the local structure of the data is effectively captured and the manifold's geometry is preserved in the lower-dimensional representation.

\subsubsection{Initialize Metric}
\label{subsec:initialize_metric}
In the process of LLE with metric learning, the initial metric \( \mathbf{M} \) can be set in several ways.
The most straightforward approach is the Euclidean initialization, where the metric \( \mathbf{M} \) is initialized to the identity matrix \( \mathbf{I} \), corresponding to the standard Euclidean distance. Mathematically, this is expressed as:
\[
\mathbf{M} = \mathbf{I}
\]
In this case, the distance between points \( \mathbf{x_i} \) and \( \mathbf{x_j} \) is equivalent to the Euclidean distance:
\[
d_\mathbf{M}(\mathbf{x_i}, \mathbf{x_j}) = \| \mathbf{x_i} - \mathbf{x_j} \|^2
\]
This initialization assumes no initial distortion or learning in the distance metric.
An alternative approach is random initialization, where \( \mathbf{M} \) is initialized as \( \mathbf{M} = \mathbf{L} \mathbf{L}^\top \), with \( \mathbf{L} \) being a randomly initialized matrix. This ensures that \( \mathbf{M} \) is PSD, a desirable property for a valid distance metric. Specifically, each element of \( \mathbf{L} \) is typically drawn from a normal distribution:
\[
L_{ij} \sim \mathcal{N}(0, \sigma^2)
\]
where \( \mathcal{N}(0, \sigma^2) \) is a normal distribution with mean 0 and variance \( \sigma^2 \). This guarantees that \( M \) remains symmetric and PSD:
\[
\mathbf{M} = \mathbf{L} \mathbf{L}^\top
\]

\begin{figure}
    \centering
    \begin{subfigure}{0.45\textwidth}
        \centering
        \includegraphics[width=\textwidth]{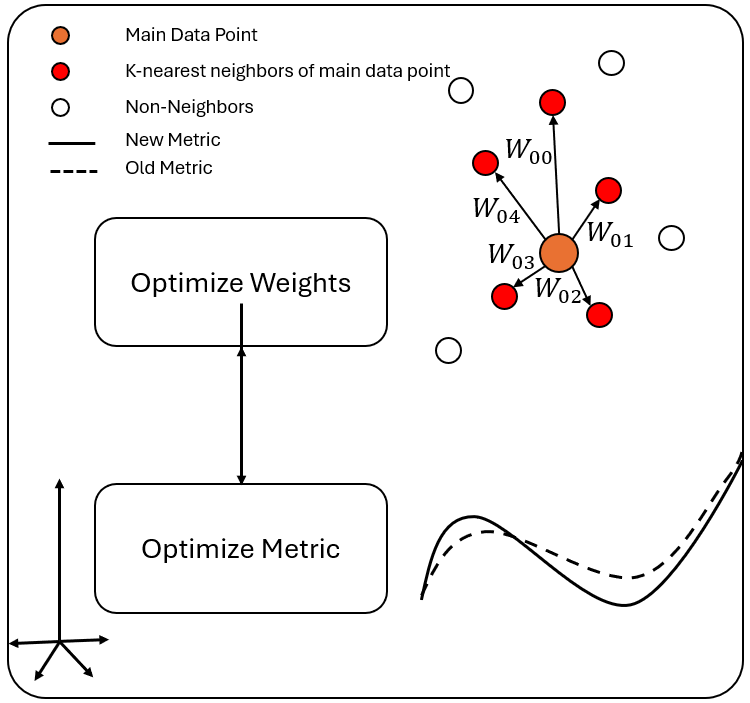}
        \caption{Reconstruction}
        \label{fig:subfig-a}
    \end{subfigure}
    \hfill
    \begin{subfigure}{0.45\textwidth}
        \centering
        \includegraphics[width=\textwidth]{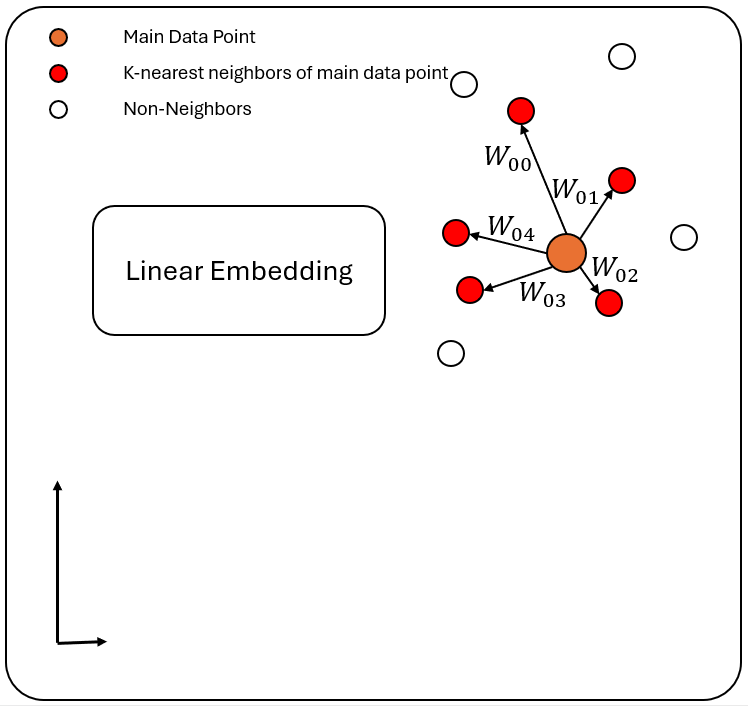}
        \caption{Linear Embedding}
        \label{fig:subfig-b}
    \end{subfigure}
    
    \caption{Optimization Problem.}
    \label{fig:optimization}
\end{figure}
\subsection{Optimization Problem}
\label{subsec:optimization_problem}
As illustrated int Fig.\ref{fig:optimization} optimization process in proposed approach consists of two key components: Reconstruction, which reconstructs each data point from its neighbors, and metric learning, which seeks to identify an optimal metric based on the reconstruction error. These two steps are iteratively repeated within a loop, and after a predefined number of epochs, the optimization process is completed.
\subsubsection{Reconstruction}
Let \(\mathbf{x}_{ij} \in \mathbb{R}^d \) denote the \(j\)-th neighbor of \(\mathbf{x}_i\), and let the matrix \( \mathbb{R}^{d \times k} \ni X_i := [\mathbf{x}_{i1}, \dots, \mathbf{x}_{ik}]\) include the \(k\) neighbors of \(\mathbf{x}_i\).
\[
\min_{{\tilde{\mathbf{W}}}} E(\mathbf{W})= \sum_{i=1}^N \left\| \mathbf{x}_i - \sum_{j \in \mathcal{N}(i)} {\mathbf{\tilde{w}}_{ij} \mathbf{x}_{ij}}\right\|_\mathbf{M}^2,
\]
after optimization, we would have optimized \( \tilde{W} \) as:
\[
\mathbf{\tilde{w}}_i = \frac{\lambda_i}{2} \mathbf{G}_i^{-1} \mathbf{1} = \frac{\mathbf{G}_i^{-1} \mathbf{1}}{\mathbf{1}^\top \mathbf{G}_i^{-1} \mathbf{1}}\]
where: 
\[ \mathbb{R}^{k \times k} \ni \mathbf{G}_i := (\mathbf{x}_i \mathbf{1}^\top - \mathbf{X}_i)^\top \mathbf{M} (\mathbf{x}_i \mathbf{1}^\top - \mathbf{X}_i)
\]\[
\mathbf{x}_i = \mathbf{x}_i \mathbf{1}^\top \mathbf{\tilde{w}}_i.
\]
\subsubsection{Metric Optimization}
We define the objective function for optimizing the metric matrix \( \mathbf{M} \) with fixed weights \( \mathbf{W} \):
\[
\min_{\mathbf{M}} E(\mathbf{M}) = \sum_{i=1}^N \left\| \mathbf{x}_i - \sum_{j \in \mathcal{N}(i)}{\mathbf{\tilde{w}}_{ij}} \mathbf{x}_j \right\|_\mathbf{M}^2.
\]
The resulting closed-form gradient descent update for \( \mathbf{M} \) is given by:
\[
\mathbf{M}^{(t+1)} = \mathbf{M}^{(t)} - \eta \sum_{i=1}^N \mathbf{r}_i \mathbf{r}_i^T,
\]
where \( \mathbf{r}_i \) represents the residual. (For derivation, refer to Appendix~\ref{app:metric_learning}).

\subsubsection{Linear Embedding}
\label{subsec:linear_embedding}
Our objective for embedding data into a lower-dimensional space with fixed weights \( \mathbf{W} \) is:
\[
\min_{\mathbf{Y}} \sum_{i=1}^N \left\| \mathbf{y}_i - \sum_{j \in \mathcal{N}(i)}{\mathbf{\tilde{w}}_{ij}} \mathbf{y}_j \right\|_2^2,
\]
subject to \( \frac{1}{n} \sum_{i=1}^n \mathbf{y}_i \mathbf{y}_i^T = \mathbf{I} \) and \( \sum_{i=1}^n \mathbf{y}_i = \mathbf{0} \).

The problem simplifies to an eigenvalue decomposition:
\[
\mathbf{M}_W \mathbf{Y} = \lambda \mathbf{Y},
\]
where \( \mathbf{M}_W = (\mathbf{I} - \mathbf{W})^T (\mathbf{I} - \mathbf{W}) \). (For detailed derivation, refer to Appendix~\ref{app:linear_embedding}).

The solution \(Y\) is constructed using the eigenvectors corresponding to the \(d\) smallest non-zero eigenvalues (excluding the smallest eigenvalue, which is zero due to the translation invariance of \(M_W\)):
\[
Y = \begin{bmatrix} v_1 & v_2 & \dots & v_{d} \end{bmatrix}^\top,
\]
where \(v_i\) is the \(i\)-th eigenvector of \(M_W\).

\subsection{Dimension-Reduced Data}
The reduced-dimensional representation of the data is:
\[
\hat{x}_i = y_i \quad \text{for } i = 1, \dots, n,
\]
where \(y_i \in \mathbb{R}^d\) is the \(i\)-th row of \(Y\).

\subsection{Theoretical Guarantee: Ensuring Positive Semi-Definiteness of \( \mathbf{M} \)}
\label{subsec:gaurantee}
In this subsection, we provide a theoretical guarantee that the matrix \( \mathbf{M} \), defined as \( \mathbf{M} = \mathbf{L}^T \mathbf{L} \), remains PSD throughout the optimization process. To achieve this, we propose an alternative approach: instead of updating \( \mathbf{M} \) directly, we update \( \mathbf{L} \), which guarantees that \( \mathbf{M} \) is always PSD.

To guarantee that \( \mathbf{M} \) remains PSD, we propose updating \( \mathbf{L} \) instead of \( \mathbf{M} \). This ensures that the product \( \mathbf{L}^T \mathbf{L} \) remains PSD throughout the optimization. The gradient of the objective function \( E(\mathbf{M}) \) with respect to \( \mathbf{L} \) is derived as follows:

\[
E(\mathbf{M}) = \sum_{i=1}^{N} \mathbf{r}_i^T \mathbf{L}^T \mathbf{L} \mathbf{r}_i = \sum_{i=1}^{N} ( \mathbf{L} \mathbf{r}_i)^T (\mathbf{L}\mathbf{r_i})
\]

Taking the derivative of \( E(\mathbf{M}) \) with respect to \( \mathbf{L} \), we get:

\[
\frac{\partial E(\mathbf{M})}{\partial \mathbf{L}} = 2 \sum_{i=1}^{N} \mathbf{L} \mathbf{r}_i \mathbf{r}_i^T
\]

This gradient expression allows us to update \( \mathbf{L} \) in the optimization process.
We update the matrix \( \mathbf{L} \) using the following gradient descent update rule:

\[
\mathbf{L}^{(t+1)} = \mathbf{L}^{(t)} - 2 \eta \sum_{i=1}^{N} \mathbf{L}^{(t)} \mathbf{r}_i \mathbf{r}_i^T
\]

where \( \eta \) is the learning rate. By updating \( \mathbf{L} \) in this manner, we guarantee that \( \mathbf{M} = \mathbf{L}^T \mathbf{L} \) remains PSD at every iteration. This is because \( \mathbf{M} \) is the product of \( \mathbf{L} \) and its transpose, and such a product is always positive semi-definite.
In some cases, to ensure that \( \mathbf{M} \) does not become singular or degenerate (i.e., with zero or near-zero eigenvalues), we can introduce a regularization term. The updated rule for \( \mathbf{L} \) with regularization becomes:

\[
\mathbf{L}^{(t+1)} = \mathbf{L}^{(t)} - 2 \eta \sum_{i=1}^{N} \mathbf{L}^{(t)} \mathbf{r}_i \mathbf{r}_i^T + \lambda \mathbf{L}^{(t)}
\]

where \( \lambda > 0 \) is a small positive constant that ensures that \( L \) remains well-conditioned. This regularization term ensures that the eigenvalues of \( M \) do not become too small, maintaining the stability of the optimization process.
The final algorithm of the proposed method is presented in Algorithm \ref{alg:alle}.

\begin{algorithm}
\caption{ALLE}
\label{alg:alle}
\begin{algorithmic}[1]
\State \textbf{Input:} Data matrix $\mathbf{X}$, number of components $n_{\text{components}}$, number of neighbors $n_{\text{neighbors}}$
\State \textbf{Output:} Lower-dimensional embedding of the data

\State Initialize the metric matrix $\mathbf{M}$.
\State Initialize $\mathbf{M}$ as the identity matrix.

\State Find $n_{\text{neighbors}}$ nearest neighbors for each data point.
\For{epoch = 1 to \texttt{max\_epochs}}
    \State Call \texttt{CholeskyDecomposition} to Decompose Metric $\mathbf{M}$ 
    \State Compute reconstruction weights using \texttt{LinearReconstruction()} with respect to the metric $\mathbf{M}$.
    \If{Optimization method = 'Adam'}
        \State Update $\mathbf{L}$ using the Adam optimizer via \texttt{AdamOptimizer()}.
    \ElsIf{Optimization method = 'SGD'}
        \State Update $\mathbf{L}$ using stochastic gradient descent (SGD) via \texttt{SGDOptimizer()}.
    \EndIf
\EndFor

\State Perform linear embedding using \texttt{LinearEmbedding()}.
\State \textbf{Return:} Lower-dimensional embedding of the data.

\end{algorithmic}
\end{algorithm}

\section{Experimental results}
In this section, we present the evaluation metrics and datasets employed in our study. Following this, we provide a comparative analysis of the results obtained from our proposed approach against those of the standard LLE and ISO LLE algorithms.

\subsection{Evaluation Metrics} \label{evaluation_metrics}
To assess the performance of the proposed Consistent LLE algorithm, we employ several evaluation metrics. These metrics capture various aspects of the algorithm's effectiveness, including the preservation of local and global data structures and the suitability of the learned low-dimensional representations for downstream tasks like clustering and classification. The following metrics are used in our evaluation:

\subsubsection{Neighborhood Preservation}
To evaluate how well the Consistent LLE preserves local neighborhood structures, we employ two standard metrics: \textit{Trustworthiness} and \textit{Continuity}, introduced by Venna and Kaski \cite{venna2001neighborhood}. 
\begin{itemize}
    \item \textbf{Trustworthiness} quantifies how many of the original nearest neighbors in high-dimensional space are retained as neighbors in the low-dimensional space. It is given by:
    \[
    T(k) = 1 - \frac{2}{n k (2n - 3k - 1)} \sum_{i=1}^{n} \sum_{j \in U(i,k)} (r(i,j) - k)
    \]
    where \( r(i,j) \) is the rank of point \( j \) in the neighborhood of point \( i \) in the original high-dimensional space, and \( U(i,k) \) is the set of points in the low-dimensional space that are not among the \( k \)-nearest neighbors of point \( i \). A higher trustworthiness score indicates better preservation of local neighborhoods.
    
    \item \textbf{Continuity} evaluates how well the low-dimensional neighbors are true neighbors in the original space. Continuity is computed as:
    \[
    C(k) = 1 - \frac{2}{n k (2n - 3k - 1)} \sum_{i=1}^{n} \sum_{j \in V(i,k)} (s(i,j) - k)
    \]
    where \( s(i,j) \) is the rank of point \( j \) in the low-dimensional space relative to point \( i \), and \( V(i,k) \) represents the neighbors in the original space that are missing in the low-dimensional embedding. Higher continuity indicates that the neighborhood structure is preserved well in both directions.
\end{itemize}

\subsubsection{Silhouette Score}
The silhouette score \cite{rousseeuw1987silhouettes} is used to evaluate the quality of clustering in the low-dimensional space. It provides a measure of how similar an object is to its own cluster compared to other clusters. The silhouette score for each data point is given by:
\[
S(i) = \frac{b(i) - a(i)}{\max(a(i), b(i))}
\]
where \( a(i) \) is the average intra-cluster distance (the average distance between the point \( i \) and other points in its cluster), and \( b(i) \) is the average nearest-cluster distance (the average distance between point \( i \) and points in the nearest cluster). The silhouette score ranges from -1 to 1, with higher values indicating better-defined clusters. We compute the overall silhouette score as the mean score across all data points.
\begin{figure}[!t]
    \centering
    \begin{subfigure}[b]{1\textwidth}
        \centering
        \includegraphics[width=\textwidth]{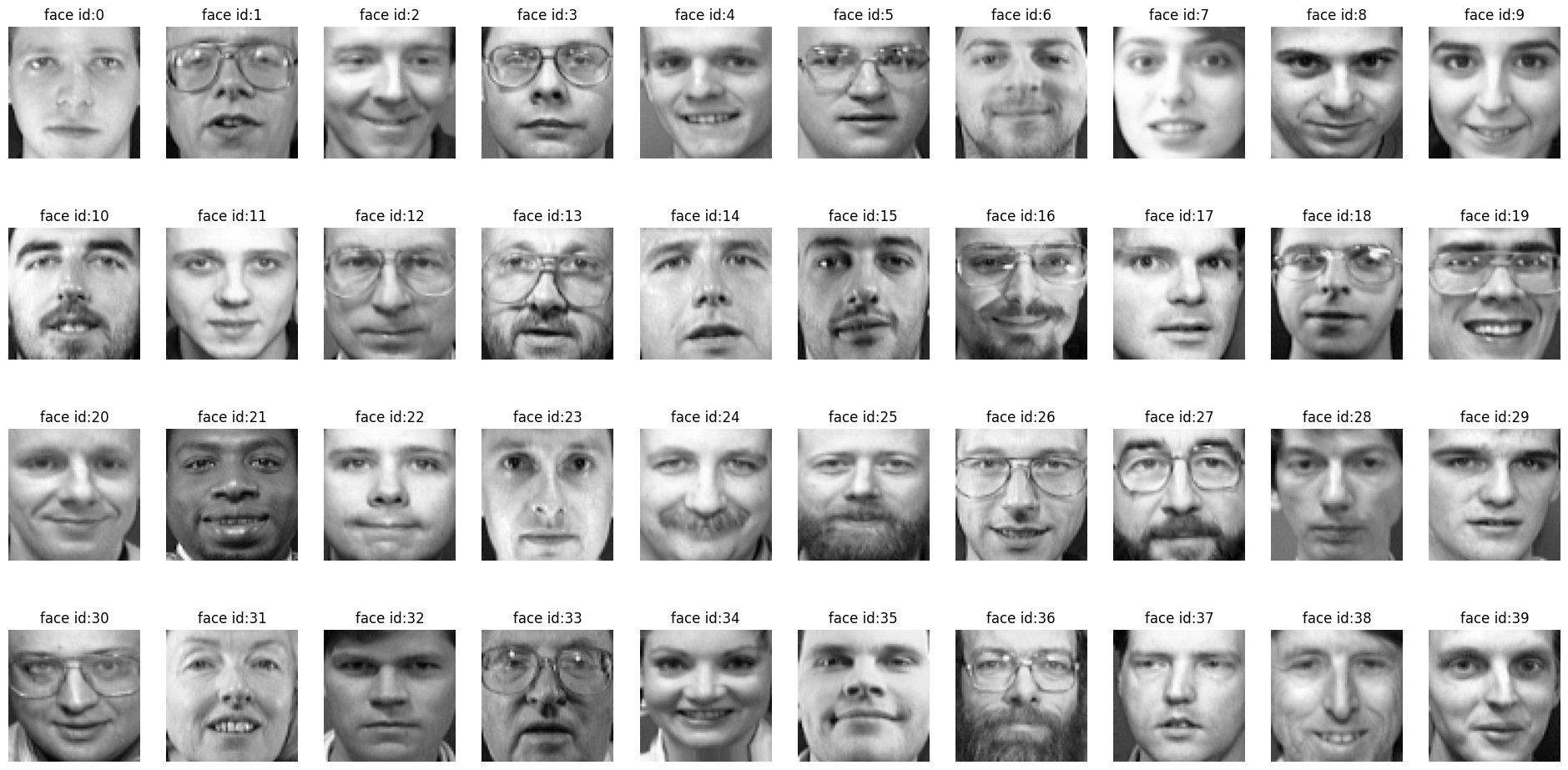}
        \caption{Olivetti}
        \label{fig:swiss_standard_lle}
    \end{subfigure}
    
    \vspace{1em} 
    
    \begin{subfigure}[b]{0.49\textwidth}
        \centering
        \includegraphics[width=0.9\textwidth]{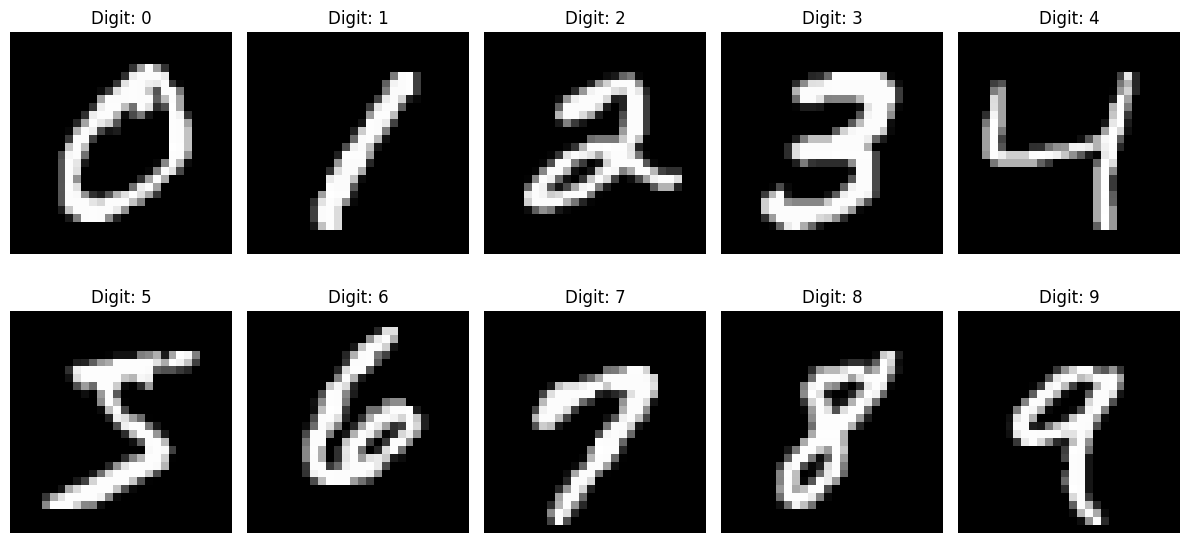}
        \caption{MNIST}
        \label{fig:mnist_standard_lle}
    \end{subfigure}
    \hfill
    \begin{subfigure}[b]{0.49\textwidth}
        \centering
        \includegraphics[width=0.9\textwidth]{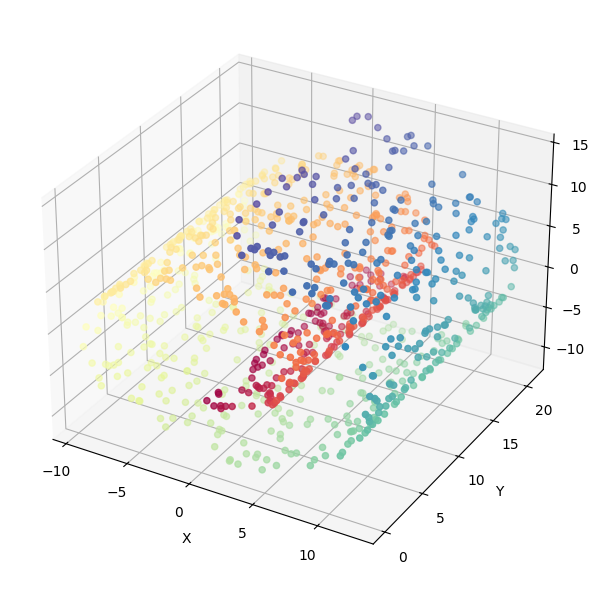}
        \caption{Swiss Roll}
        \label{fig:mnist_metric_lle}
    \end{subfigure}
    
    \caption{Visualizations of three datasets: (a) Olivetti Faces, (b) MNIST, and (c) Swiss Roll.}
    \label{fig:datasets}
\end{figure}

\subsubsection{Accuracy Score}
When the dataset has class labels, we assess the accuracy of the learned embedding in a classification task. After performing dimensionality reduction using Consistent LLE, we train a classifier (e.g., k-Nearest Neighbors) on the low-dimensional representations and compute the classification accuracy. The accuracy is defined as:
\[
\text{Accuracy} = \frac{\text{Number of Correct Predictions}}{\text{Total Number of Predictions}}
\]
This metric helps us evaluate whether the low-dimensional embedding preserves class separability, which is essential for downstream classification tasks.

\subsection{Datasets and Experimental Setup}
In our experiments, we employed the Adam optimizer \cite{kingma2014adam}. Additionally, we evaluated the performance of the proposed ALLE algorithm using a range of well-established datasets. These datasets, encompassing both synthetic and real-world data, are widely recognized in the dimensionality reduction literature and serve as standard benchmarks for algorithm evaluation. 

\subsubsection{MNIST}
The MNIST dataset \cite{deng2012mnist}, consisting of 70,000 grayscale images of handwritten digits (28x28 pixels), serves as a benchmark for classification and dimensionality reduction. For this study, we evaluated the algorithm on two variations: the full dataset with 10 digit classes (0–9), offering high variability and challenging class separability, and a subset with 6 digit classes (0–5), providing a simpler task with reduced inter-class variability while maintaining the need for clear separability.

\subsubsection{Swiss Roll}
The Swiss Roll dataset is a popular synthetic dataset used for testing manifold learning algorithms. It consists of points uniformly sampled from a two-dimensional manifold embedded in three-dimensional space. The dataset is challenging because it requires the algorithm to unfold the non-linear manifold into a lower-dimensional space while preserving its intrinsic structure.

\subsubsection{Olivetti Faces}
The Olivetti Faces Dataset consists of 400 grayscale images of size 64 × 64 pixels, representing 40 distinct classes (individuals) with 10 images per class. The images capture variations in facial expressions and poses under controlled conditions, making it a widely used benchmark for tasks such as dimensionality reduction, face recognition, and clustering. 

\begin{table}[t]
\centering
\renewcommand{\arraystretch}{1.5} 
\begin{tabular}
{|l|c c c|c c c|}\hline
\multirow{2}{*}{\textbf{Data set}} & \multicolumn{3}{|c|}{\textbf{Trustworthy}} & \multicolumn{3}{|c|}{\textbf{Continuity}} \\ \cline{2-7}
 & \textbf{LLE} & \textbf{ISO LLE}  & \textbf{ALLE} & \textbf{LLE} & \textbf{ISO LLE} & \textbf{ALLE} \\ \hline
\text{Swiss roll} & \textbf{0.9967} & 0.9872 & 0.9929 & \textbf{0.9967} & 0.9768 & 0.9880 \\ \hline
\text{Scaled swiss roll} & 0.9927 & 0.9872 & \textbf{0.9954} & \textbf{0.9982} & 0.9768 & 0.9954 \\ \hline
\text{MNIST} & 0.9704 & 0.9683 & \textbf{0.9752} & 0.9248 & 0.9057 & \textbf{0.9322} \\ \hline
\text{Iris} & \textbf{0.9669} & 0.9592 & 0.9559 & \textbf{0.9563} & 0.9274 & 0.9294 \\ \hline
\text{Olivetti(40 classes)}& 0.8966& \textbf{0.9359}& 0.8321& 0.8863 & \textbf{0.8949} & 0.8267 \\ \hline
\text{Olivetti(4 classes)}& 0.9158 & \textbf{0.9298} & 0.9222 & 0.9476 & \textbf{0.9560} & 0.9473 \\ \hline
\end{tabular}
\caption{\label{table:1}Comparison of Standard LLE, ISO LLE and ALLE on various datasets based on Trustworthy and Continuity metrics.}
\end{table}

\begin{table}[t]
\centering
\scriptsize 
\renewcommand{\arraystretch}{1.5} 
\setlength{\tabcolsep}{3pt}
\small 
\begin{tabular}{|l|c c c|c c c|c c c|} \hline
\multirow{2}{*}{\textbf{Data set}} & \multicolumn{3}{|c|}{\textbf{SVC}} & \multicolumn{3}{|c|}{\textbf{KNN}} & \multicolumn{3}{|c|}{\textbf{Silhouette Score}} \\ \cline{2-10}
 & \textbf{LLE} & \textbf{ISO LLE}  & \textbf{ALLE} & \textbf{LLE} & \textbf{ISO LLE}  & \textbf{ALLE} & \textbf{LLE} & \textbf{ISO LLE}  & \textbf{ALLE} \\ \hline
\text{MNIST} & 0.8797 & 0.8525 & \textbf{0.9053} & 0.9382 & 0.9126 & \textbf{0.9449} & \textbf{0.6346} & 0.6225 & 0.4922 \\ \hline
\text{Iris} & 0.9133 & 0.88 & \textbf{0.9533} & 0.9466 & 0.92 & \textbf{0.9533} & \textbf{0.7151} & 0.6887 & 0.7035 \\ \hline
\text{Olivetti(40 classes)} & 0.5675 & 0.44 & \textbf{0.725} & 0.8175 & 0.6675 & \textbf{0.85} & \textbf{0.4865} & 0.2923 & 0.2525 \\ \hline 
\text{Olivetti(4 classes)} & 0.825 & 0.775 & \textbf{0.95} & 0.925 & 0.875 & \textbf{0.95} & 0.6440 & 0.8833 & \textbf{0.6556} \\ \hline 
\end{tabular}
\caption{\label{table:2}Comparison of Standard LLE, ISO LLE and ALLE on various datasets based on SVC, KNN, and Silhouette Score metrics.}
\end{table}

\begin{figure}[!t]
    \centering
    \begin{subfigure}[b]{.45\textwidth}
        \centering
        \includegraphics[width=\textwidth]{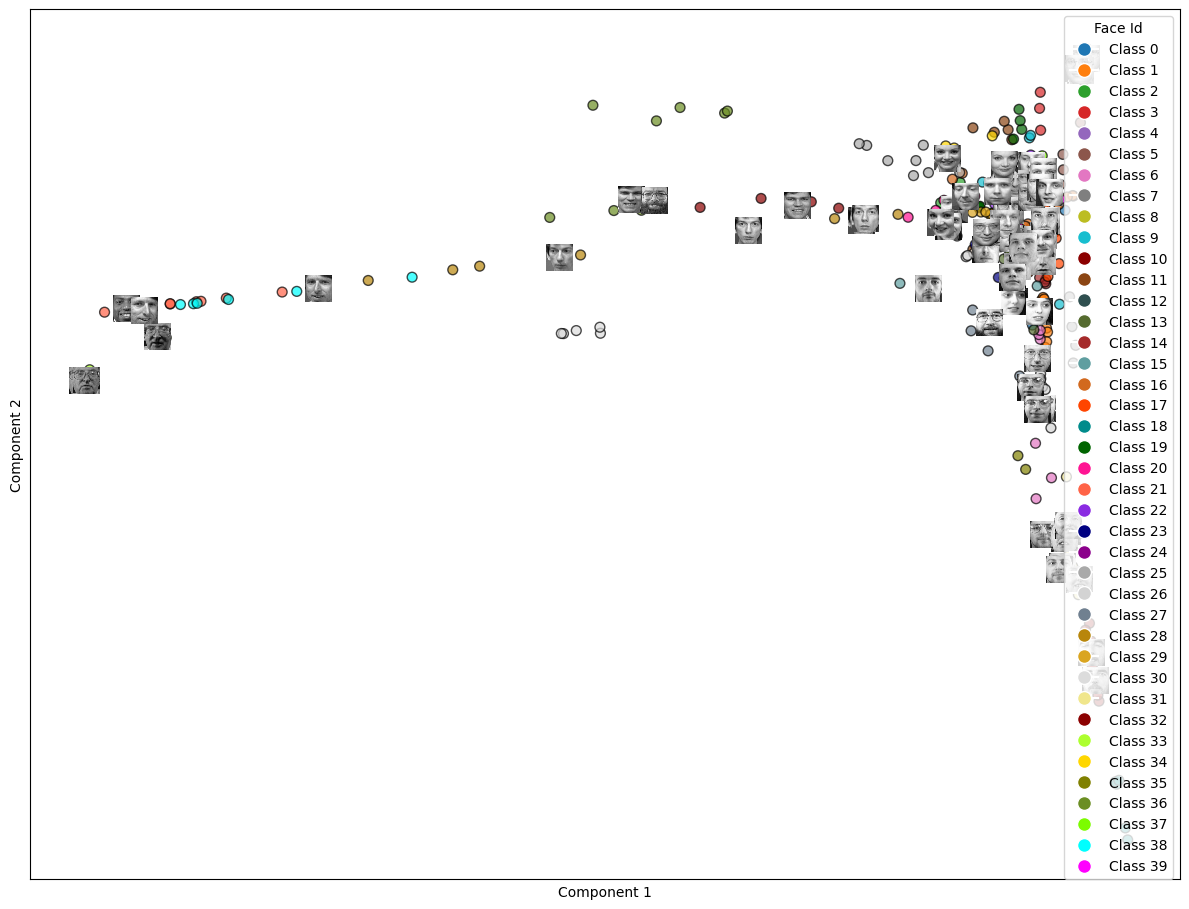}
        \caption{Olivetti using Standard LLE}
        \label{fig:comparsion-slle_olivetti}
    \end{subfigure}
    \hfill
    \begin{subfigure}[b]{.45\textwidth}
        \centering
        \includegraphics[width=\textwidth]{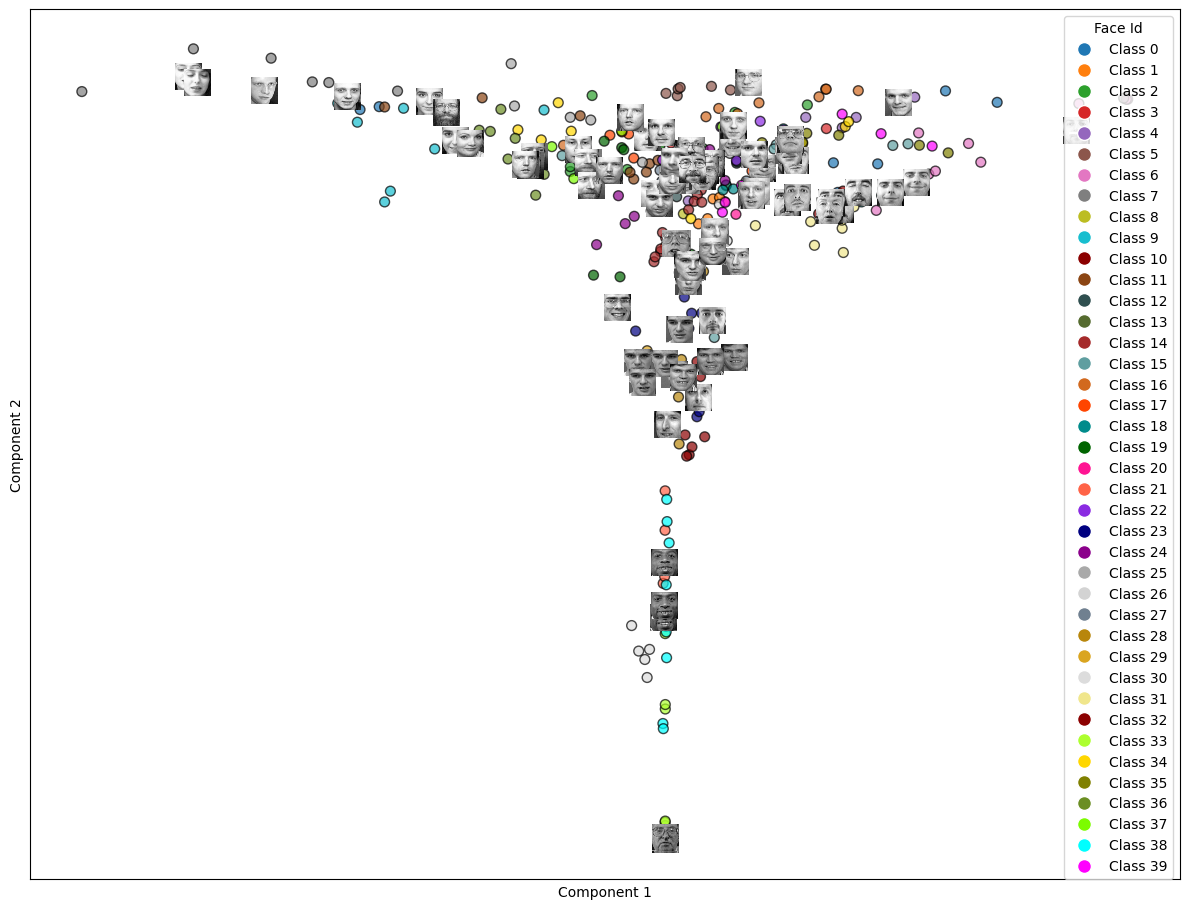}
        \caption{Olivetti Classes using ALLE}
        \label{fig:comparsion-alle_olivetti}
    \end{subfigure}
    \begin{subfigure}[b]{0.45\textwidth}
        \centering
        \includegraphics[width=\textwidth]{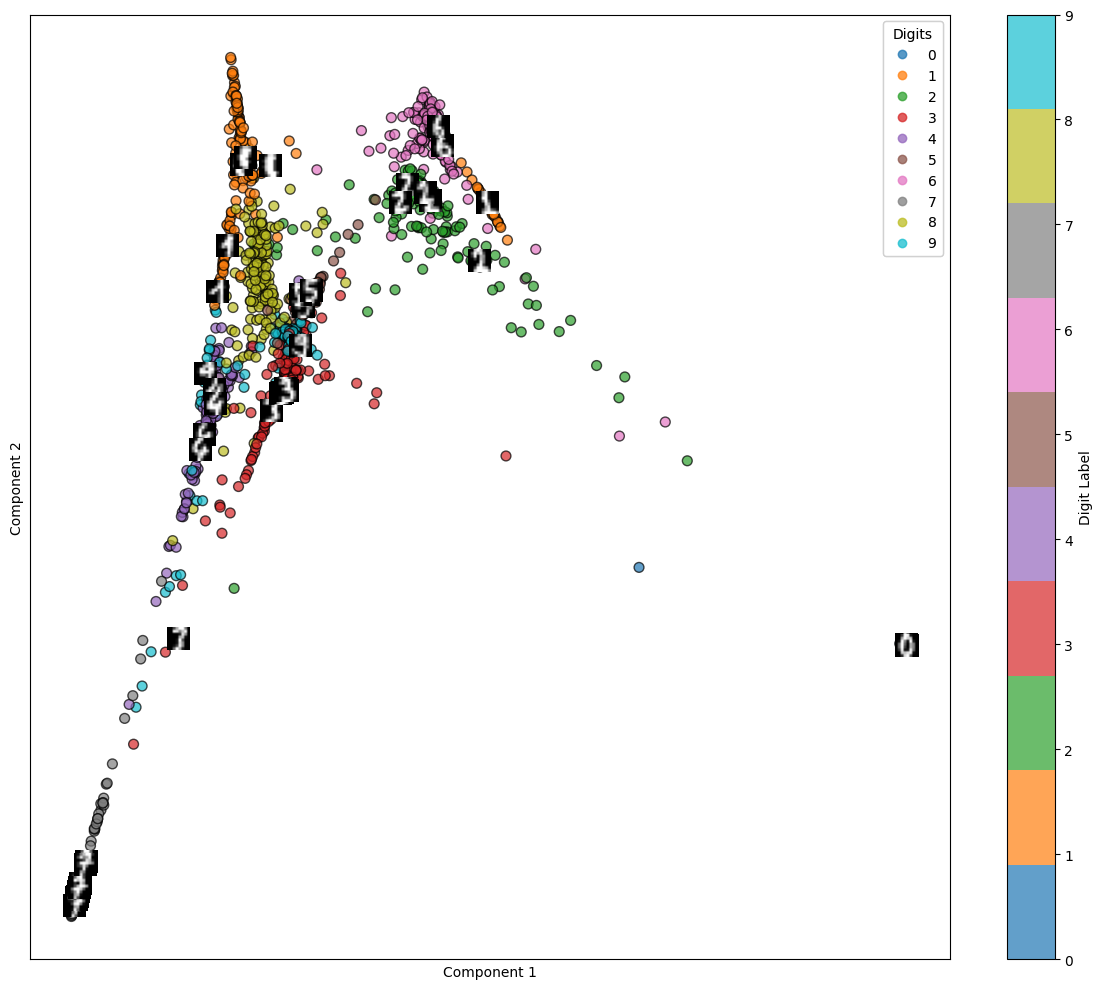}
        \caption{MNIST 10 Classes using Standard LLE}
        \label{fig:comparsion-slle_mnist}
    \end{subfigure}
    \hfill
    \begin{subfigure}[b]{0.45\textwidth}
        \centering
        \includegraphics[width=\textwidth]{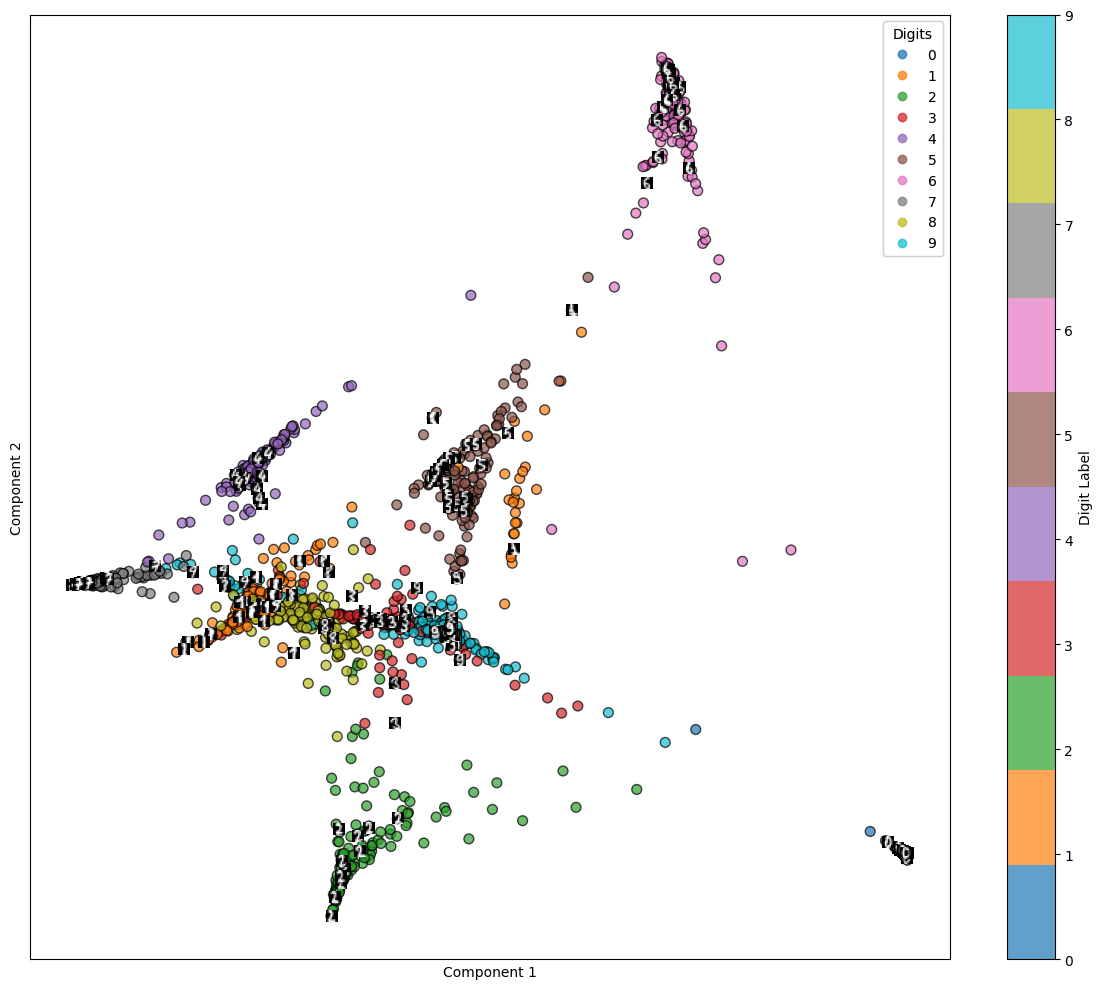}
        \caption{MNIST 10 Classes ALLE}
        \label{fig:comparsion-alle_mnist}
    \end{subfigure}
    
    \vskip\baselineskip
    
    \begin{subfigure}[b]{0.45\textwidth}
        \centering
        \includegraphics[width=\textwidth]{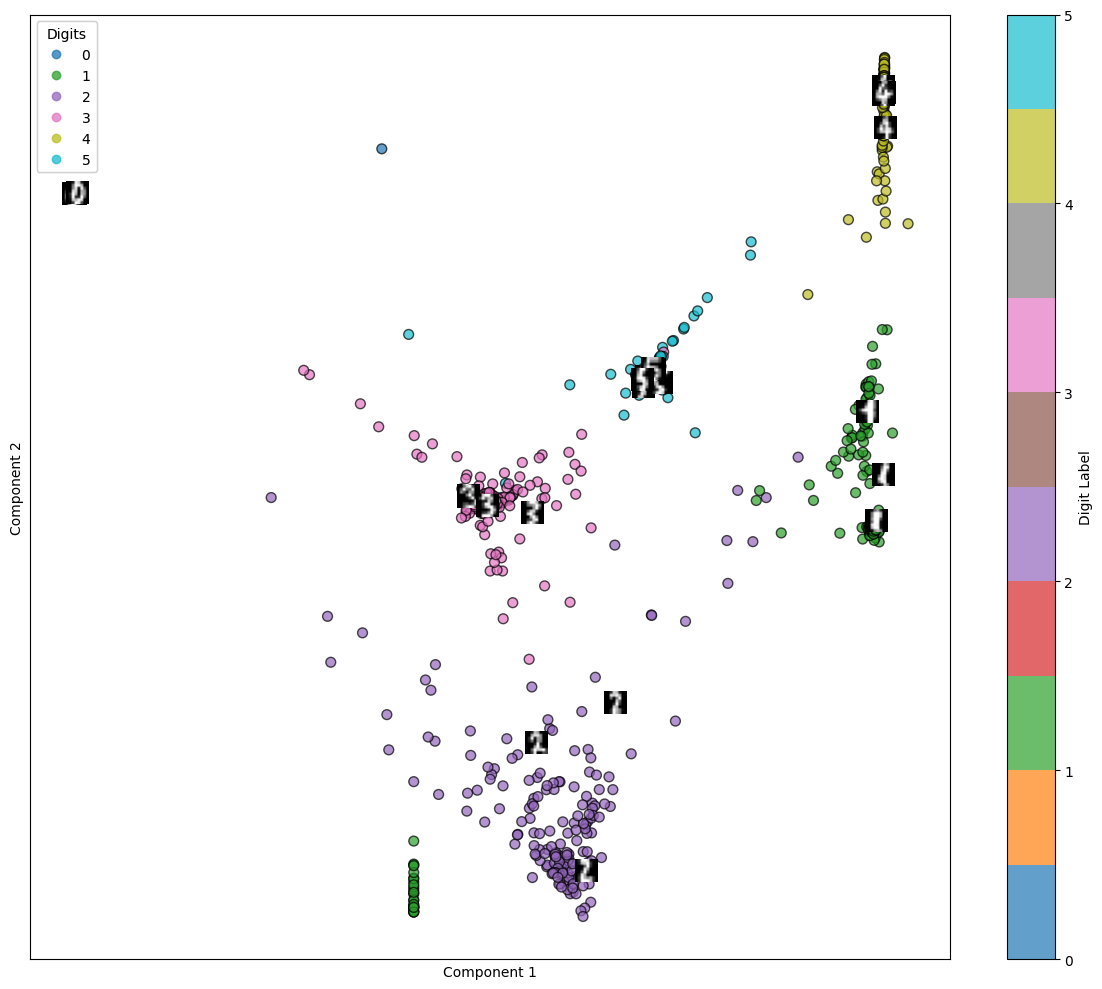}
        \caption{MNIST 6 Classes using Standard LLE}
        \label{fig:comparsion-slle_mnist6}
    \end{subfigure}
    \hfill
    \begin{subfigure}[b]{0.45\textwidth}
        \centering
        \includegraphics[width=\textwidth]{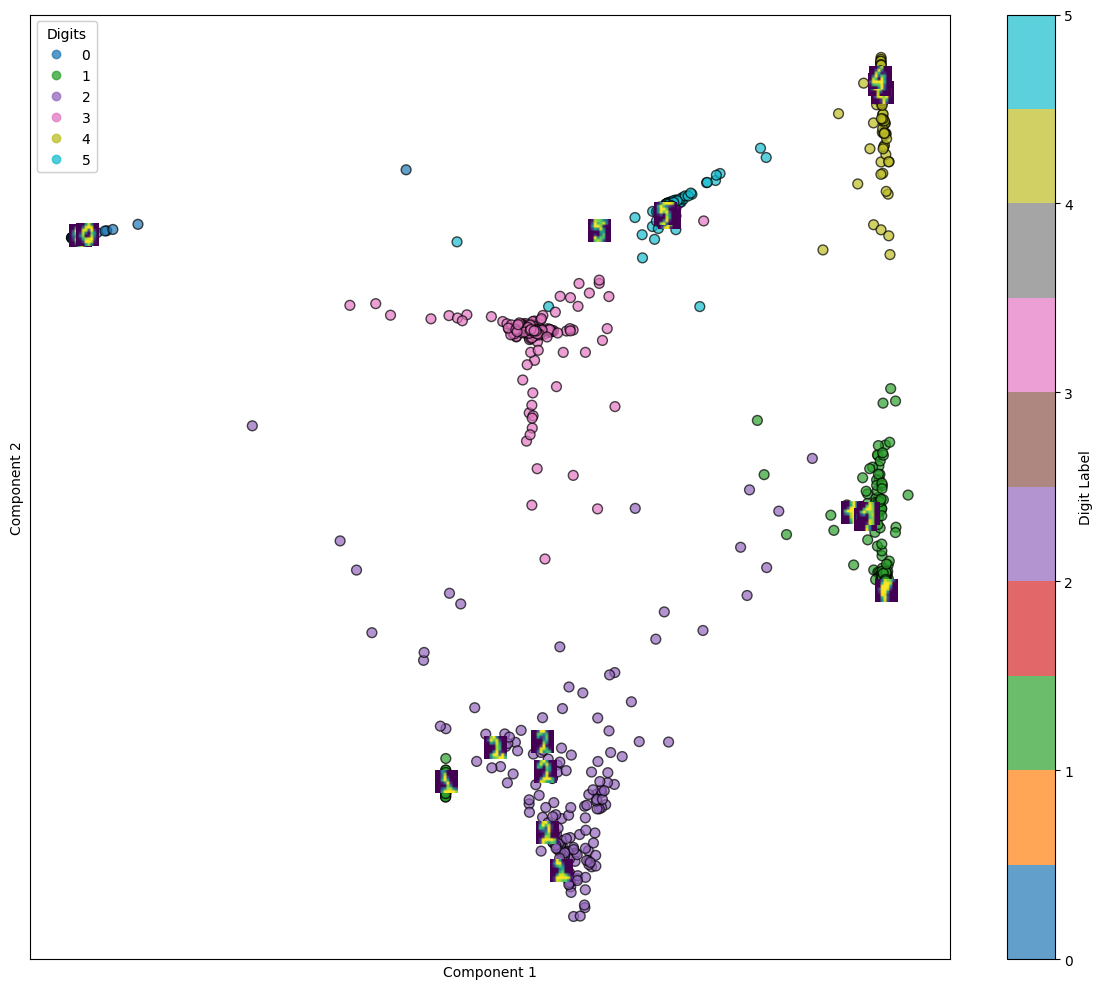}
        \caption{MNIST 6 Classes using ALLE}
        \label{fig:comparsion-alle_mnist6}
    \end{subfigure}
    \caption{\label{fig:comparsion}Results on Three Datasets Using Standard LLE and ALLE}
\end{figure}

\subsection{Compare Results}

In this section, we provide a comprehensive comparative analysis of the proposed ALLE algorithm against the standard LLE approach. The evaluation metrics utilized in this comparison are introduced in Section \ref{evaluation_metrics}. 

The results clearly demonstrate the effectiveness of the adaptive metric in preserving neighborhood structures and improving the quality of low-dimensional embeddings. As illustrated in Fig. \ref{fig:comparsion}, the differences between Fig. \ref{fig:comparsion-slle_olivetti} and Fig. \ref{fig:comparsion-alle_olivetti} highlight the capability of the ALLE algorithm to adaptively stretch and compress data in multiple directions. This flexibility leads to significant improvements in trustworthiness and continuity, as evidenced in Table \ref{table:1}. Furthermore, the proposed algorithm enhances classification accuracy when evaluated using the Support Vector Classifier (SVC) algorithm, as detailed in Table \ref{table:2}. Notably, while the KNN score for standard LLE shows slightly better results, this can be attributed to the simpler geometric structure of its output, which relies purely on Euclidean distance.

Similarly, Fig. \ref{fig:comparsion-slle_mnist} and Fig. \ref{fig:comparsion-alle_mnist} underscore the advantages of leveraging a learned adaptive metric for bending the topological space. This approach yields superior visualizations in lower dimensions, particularly for complex datasets. For instance, the inter-class and intra-class relationships between digits such as ``2'' and ``6'' and ``3'' and ``9'' in the MNIST dataset are more distinctly represented. Table \ref{table:2} further corroborates these observations, demonstrating improved silhouette scores under the ALLE algorithm. Additionally, classification accuracy using the SVC algorithm is notably enhanced for both the standard MNIST dataset and the reduced MNIST dataset with 10 classes.

As shown in Table \ref{table:1}, the continuity and trustworthiness of the Swiss roll and scaled Swiss roll datasets have shown slight improvement. The results indicate that the visualization of ALLE is indistinguishable from that of standard LLE on this dataset. This outcome can be attributed to the sufficiency of the Euclidean distance in the linear reconstruction process for this dataset, rendering the metric modifications inconsequential.

Interestingly, while standard LLE achieves better continuity scores for specific datasets, this outcome can be attributed to the bending of the topological space by ALLE. This adaptation creates a clearer separation between data points from different classes, which, under the initial KNN-based neighbor selection, may otherwise have been incorrectly classified as neighbors. 

These findings confirm that the ALLE algorithm effectively balances the trade-off between improving global visualization quality and maintaining local neighborhood structures, making it a powerful tool for dimensionality reduction and data analysis.

\subsection{Conclusion and Future Work}

To wrap up, the ALLE algorithm has shown clear advantages over the standard LLE approach. It preserves neighborhood structures more effectively, creates higher-quality low-dimensional embeddings, and improves classification accuracy. By adapting the metric to stretch and compress the data, ALLE opens new possibilities for analyzing and visualizing complex datasets. However, some trade-offs, such as slightly lower continuity scores in certain cases, provide areas for further refinement.

Looking ahead, there are exciting opportunities to build on this work. Future efforts could focus on improving the optimization process for the adaptive metric to strike an even better balance between trustworthiness and continuity. Scaling the algorithm for larger datasets and integrating it with deep learning frameworks could make it even more powerful and versatile. Exploring real-time applications where the adaptive metric evolves dynamically with the data is another promising avenue. These steps will help solidify ALLE as a method for dimensionality reduction and visualization in increasingly complex data scenarios.

\clearpage
\appendix
\appendix
\section{Proofs and Derivations}

\subsection{Metric Optimization}
\label{app:metric_learning}
Here, we provide the detailed derivation of the objective function for metric learning. Starting from the objective function:
\[
E(\mathbf{M}) = \sum_{i=1}^N \left\| \mathbf{x}_i - \sum_{j \in \mathcal{N}(i)}{\mathbf{\tilde{w}}_{ij}} \mathbf{x}_j \right\|_\mathbf{M}^2,
\]
expanding this, we have:
\[
E(\mathbf{M}) = \sum_{i=1}^N {(\mathbf{x}_i - \sum_{j \in \mathcal{N}(i)} \mathbf{\tilde{w}}_{ij} \mathbf{x}_j)}^T \mathbf{M} {(\mathbf{x}_i - \sum_{j \in \mathcal{N}(i)} \mathbf{\tilde{w}}_{ij} \mathbf{x}_j)}.
\]
Defining \( \mathbf{r}_i = \mathbf{x}_i - \sum_{j \in \mathcal{N}(i)} \mathbf{\tilde{w}}_{ij} \mathbf{x}_j \), we substitute to get:
\[
E(\mathbf{M}) = \sum_{i=1}^N \mathbf{r}_i^T \mathbf{M} \mathbf{r}_i.
\]

Then, differentiating \( E(\mathbf{M}) \) with respect to \( \mathbf{M} \) gives:
\[
\frac{\partial E(\mathbf{M})}{\partial \mathbf{M}} = \sum_{i=1}^N \mathbf{r}_i \mathbf{r}_i^T.
\]
This result allows us to update \( \mathbf{M}^{(t+1)} \) using gradient descent:
\[
\mathbf{M}^{(t+1)} = \mathbf{M}^{(t)} - \eta \sum_{i=1}^N \mathbf{r}_i \mathbf{r}_i^T.
\]

\subsection{Linear Embedding}
\label{app:linear_embedding}
The linear embedding optimization problem is formulated as:
\[
\min_{\mathbf{Y}} \sum_{i=1}^N \left\| \mathbf{y}_i - \sum_{j \in \mathcal{N}(i)}{\mathbf{\tilde{w}}_{ij}} \mathbf{y}_j \right\|_2^2,
\]
with the constraints:
\[
\frac{1}{n} \sum_{i=1}^n \mathbf{y}_i \mathbf{y}_i^T = \mathbf{I}, \quad \sum_{i=1}^n \mathbf{y}_i = \mathbf{0}.
\]

Expanding the objective function, we have:
\[
\min_{\mathbf{Y}} \sum_{i=1}^N \left( \mathbf{y}_i - \sum_{j=1}^{n} \mathbf{w}_{ij} \mathbf{y}_j \right)^T \left( \mathbf{y}_i - \sum_{j=1}^{n} \mathbf{w}_{ij} \mathbf{y}_j \right).
\]
In matrix form, this becomes:
\[
\min_{\mathbf{Y}} \, \text{tr}\left( \mathbf{Y}^T (\mathbf{I} - \mathbf{W})^T (\mathbf{I} - \mathbf{W}) \mathbf{Y} \right).
\]
Defining \( \mathbf{M}_W = (\mathbf{I} - \mathbf{W})^T (\mathbf{I} - \mathbf{W}) \), we can simplify to:
\[
\min_{\mathbf{Y}} \, \text{tr}\left( \mathbf{Y}^T \mathbf{M}_W \mathbf{Y} \right).
\]

To solve this using Lagrange multipliers, define the Lagrangian:
\[
L(\mathbf{Y}, \Lambda) = \text{tr}(\mathbf{Y}^T \mathbf{M}_W \mathbf{Y}) - \text{tr}\left( \Lambda \left( \frac{1}{n} \mathbf{Y}^T \mathbf{Y} - \mathbf{I} \right) \right).
\]
Taking the derivative with respect to \( \mathbf{Y} \):
\[
\frac{\partial L}{\partial \mathbf{Y}} = 2 \mathbf{M}_W \mathbf{Y} - 2 \frac{1}{n} \mathbf{Y} \Lambda = 0.
\]
This leads to an eigenvalue problem:
\[
\mathbf{M}_W \mathbf{Y} = \frac{1}{n} \mathbf{Y} \Lambda,
\]
or equivalently:
\[
\mathbf{M}_W \mathbf{Y} = \lambda \mathbf{Y}.
\]

\bibliography{refrence}

\end{document}


\maketitle

\section{Appendix}

\subsection{Convergence Condition}

For the iterative update of \( \mathbf{M} \):

\[
\mathbf{M}^{(t+1)} = \mathbf{M}^{(t)} - \eta \sum_{i=1}^N \mathbf{r}_i \mathbf{r}_i^T,
\]

to lead to convergence, the following condition must hold:

\[
E(\mathbf{M}^{(t+1)}) < E(\mathbf{M}^{(t)}) \quad \text{for all iterations } t.
\]

\textbf{Choice of Learning Rate \( \eta \)}

To ensure that the learning rate \( \eta \) does not cause divergence and allows the algorithm to converge to a local minimum, we need to choose \( \eta \) such that:

\[
0 < \eta < \frac{2}{\lambda_{\max}},
\]

where \( \lambda_{\max} \) is the maximum eigenvalue of the Hessian of the error function \( E(\mathbf{M}) \).

\textbf{Justification of the Learning Rate Condition:}

- **Hessian Matrix**: The Hessian matrix \( H \) of the error function \( E(\mathbf{M}) \) is defined as:

  \[
  H = \frac{\partial^2 E(\mathbf{M})}{\partial \mathbf{M}^2} = \sum_{i=1}^N \mathbf{r}_i \mathbf{r}_i^T,
  \]

  which is positive semi-definite if \( \mathbf{r}_i \) are not all zero.

- **Eigenvalues and Stability**: The convergence criterion is derived from analyzing the quadratic form associated with the Hessian. If \( \eta \) is too large, the update may overshoot, causing \( E(\mathbf{M}) \) to increase rather than decrease. 

- **Sufficient Condition for Descent**: The condition \( 0 < \eta < \frac{2}{\lambda_{\max}} \) ensures that the updates remain stable and maintain a descent direction. Specifically, using Taylor expansion, we have:

  \[
  E(\mathbf{M}^{(t+1)}) \approx E(\mathbf{M}^{(t)}) - \eta \mathbf{g}^T \mathbf{g} + \frac{1}{2} \eta^2 \mathbf{g}^T H \mathbf{g},
  \]

  where \( \mathbf{g} = \sum_{i=1}^N \mathbf{r}_i \mathbf{r}_i^T \) is the gradient. The term \( \eta^2 \mathbf{g}^T H \mathbf{g} \) becomes significant for larger \( \eta \). Thus, bounding \( \eta \) by \( \frac{2}{\lambda_{\max}} \) ensures that the second-order term does not dominate and the overall error decreases.

### Conclusion

For convergence of the iterative updates in your formulation, it is crucial to choose the learning rate \( \eta \) within the bounds:

\[
0 < \eta < \frac{2}{\lambda_{\max}}.
\]

This ensures that the updates will consistently reduce the error \( E(\mathbf{M}) \) over iterations, leading to convergence towards a local minimum.

To analyze the convergence of the iterative update for \(\mathbf{M}\) in detail, we start from the error function \(E(\mathbf{M})\) and its gradient, leading us to determine the appropriate bounds for the learning rate \(\eta\).

### Error Function and Gradient

The error function is defined as:

\[
E(\mathbf{M}) = \sum_{i=1}^N \mathbf{r}_i^T \mathbf{M} \mathbf{r}_i.
\]

The gradient with respect to \(\mathbf{M}\) is given by:

\[
\frac{\partial E(\mathbf{M})}{\partial \mathbf{M}} = \sum_{i=1}^N \mathbf{r}_i \mathbf{r}_i^T.
\]

### Update Rule

Using gradient descent, we update \(\mathbf{M}\) as follows:

\[
\mathbf{M}^{(t+1)} = \mathbf{M}^{(t)} - \eta \sum_{i=1}^N \mathbf{r}_i \mathbf{r}_i^T.
\]

### Taylor Expansion for Analysis

To analyze how \(E(\mathbf{M})\) changes, we can use a Taylor expansion around \(\mathbf{M}^{(t)}\):

\[
E(\mathbf{M}^{(t+1)}) \approx E(\mathbf{M}^{(t)}) + \nabla E(\mathbf{M}^{(t)})^T (\mathbf{M}^{(t+1)} - \mathbf{M}^{(t)}) + \frac{1}{2} (\mathbf{M}^{(t+1)} - \mathbf{M}^{(t)})^T H (\mathbf{M}^{(t+1)} - \mathbf{M}^{(t)}),
\]

where \(H\) is the Hessian of \(E(\mathbf{M})\) evaluated at \(\mathbf{M}^{(t)}\):

\[
H = \sum_{i=1}^N \mathbf{r}_i \mathbf{r}_i^T.
\]

Substituting \(\mathbf{M}^{(t+1)} - \mathbf{M}^{(t)} = -\eta \sum_{i=1}^N \mathbf{r}_i \mathbf{r}_i^T\):

\[
E(\mathbf{M}^{(t+1)}) \approx E(\mathbf{M}^{(t)}) - \eta \nabla E(\mathbf{M}^{(t)})^T \sum_{i=1}^N \mathbf{r}_i \mathbf{r}_i^T + \frac{1}{2} \left(-\eta \sum_{i=1}^N \mathbf{r}_i \mathbf{r}_i^T\right)^T H \left(-\eta \sum_{i=1}^N \mathbf{r}_i \mathbf{r}_i^T\right).
\]

### Simplifying the Terms

The first-order term simplifies as:

\[
-\eta \nabla E(\mathbf{M}^{(t)})^T \sum_{i=1}^N \mathbf{r}_i \mathbf{r}_i^T = -\eta \sum_{i=1}^N \mathbf{r}_i^T \mathbf{M}^{(t)} \mathbf{r}_i,
\]

and the second-order term simplifies to:

\[
\frac{\eta^2}{2} \left(\sum_{i=1}^N \mathbf{r}_i \mathbf{r}_i^T\right)^T H \left(\sum_{i=1}^N \mathbf{r}_i \mathbf{r}_i^T\right).
\]

### Condition for Decrease in Error

To ensure that \(E(\mathbf{M}^{(t+1)}) < E(\mathbf{M}^{(t)})\), we require:

\[
-\eta \sum_{i=1}^N \mathbf{r}_i^T \mathbf{M}^{(t)} \mathbf{r}_i + \frac{\eta^2}{2} \left(\sum_{i=1}^N \mathbf{r}_i \mathbf{r}_i^T\right)^T H \left(\sum_{i=1}^N \mathbf{r}_i \mathbf{r}_i^T\right) < 0.
\]

### Analyzing the Conditions

1. **First-Order Term**: For the first term, if \(\eta\) is positive and \(\nabla E(\mathbf{M}^{(t)})\) is negative (which it should be for a minimization problem), this term decreases the error.

2. **Second-Order Term**: The second term must not be too large. We want:

   \[
   \frac{\eta^2}{2} \left(\sum_{i=1}^N \mathbf{r}_i \mathbf{r}_i^T\right)^T H \left(\sum_{i=1}^N \mathbf{r}_i \mathbf{r}_i^T\right) < \eta \sum_{i=1}^N \mathbf{r}_i^T \mathbf{M}^{(t)} \mathbf{r}_i.
   \]

3. **Bounding \( \eta \)**: By analyzing the terms, we can derive:

   - Let \( \lambda_{\max} \) be the largest eigenvalue of the Hessian \( H \):
   \[
   \left(\sum_{i=1}^N \mathbf{r}_i \mathbf{r}_i^T\right)^T H \left(\sum_{i=1}^N \mathbf{r}_i \mathbf{r}_i^T\right) \leq \lambda_{\max} \left\|\sum_{i=1}^N \mathbf{r}_i \mathbf{r}_i^T\right\|^2.
   \]

   Then, we can say that if we want:

   \[
   \frac{\eta^2}{2} \lambda_{\max} \left\|\sum_{i=1}^N \mathbf{r}_i \mathbf{r}_i^T\right\|^2 < \eta \sum_{i=1}^N \mathbf{r}_i^T \mathbf{M}^{(t)} \mathbf{r}_i,
   \]

   we can rearrange this to find a suitable range for \(\eta\):

   \[
   \eta < \frac{2 \sum_{i=1}^N \mathbf{r}_i^T \mathbf{M}^{(t)} \mathbf{r}_i}{\lambda_{\max} \left\|\sum_{i=1}^N \mathbf{r}_i \mathbf{r}_i^T\right\|^2}.
   \]

### Final Condition

To ensure convergence, we can summarize:

\[
0 < \eta < \frac{2}{\lambda_{\max}},
\]

This condition guarantees that the updates lead to a decrease in the error function, avoiding overshooting and ensuring that the second-order term does not dominate, thereby allowing the algorithm to converge toward a minimum of \(E(\mathbf{M})\).

\bibliography{refrence}